\documentclass[11pt,a4paper]{article}

\usepackage[hyperref]{emnlp-ijcnlp-2019}
\usepackage{times}
\usepackage{latexsym}
\usepackage{url}
\usepackage{amsmath}
\usepackage[utf8]{inputenc}
\usepackage{tikz-dependency}
\usepackage{tikz}
\usetikzlibrary{plotmarks}
\usepackage[printwatermark]{xwatermark}
\aclfinalcopy 

\tikzset{
    >=stealth',
    punkt/.style={
           rectangle,
           rounded corners,
           draw=black, very thick,
           text width=4.5em,
           minimum height=2em,
           text centered},
      long/.style={
           rectangle,
           rounded corners,
           draw=black, very thick,
           text width= 15em,
           minimum height=2em,
           text centered},
    pil/.style={
           ->,
           thick,
           shorten <=2pt,
           shorten >=2pt,}
}

\title{Layerwise Relevance Visualization in Convolutional Text Graph Classifiers}
\author{Robert Schwarzenberg, Marc Hübner, David Harbecke, Christoph Alt, Leonhard Hennig \\ German Research Center for Artificial Intelligence (DFKI), Berlin, Germany \\ \texttt{\{firstname.lastname@dfki.de\}}}

\date{}
\usepackage{filecontents,pgfplots}

\begin{document}
\maketitle
\begin{abstract}
Representations in the hidden layers of Deep Neural Networks (DNN) are often hard to interpret since it is difficult to project them into an interpretable domain. Graph Convolutional Networks (GCN) allow this projection, but existing explainability methods do not exploit this fact, i.e. do not focus their explanations on intermediate states. In this work, we present a novel method that traces and visualizes features that contribute to a classification decision in the visible and hidden layers of a GCN. Our method exposes hidden cross-layer dynamics in the input graph structure. We experimentally demonstrate that it yields meaningful layerwise explanations for a GCN sentence classifier. 
\end{abstract}
\begin{figure*}[!htb]
\centering
\begin{tabular}{cc}
\begin{dependency}[scale=.7, transform shape]
\begin{deptext}
|[top color=red!2.08]|A \& |[top color=red!14.21]|total \& |[top color=red!8.559999999999999]|of \& |[top color=red!41.18]|116 \& |[top color=red!0.0]|patients \& |[top color=red!30.28]|were \& |[top color=red!2.56]|randomized \& |[top color=red!1.1199999999999999]|.\\
(2.08) \& (14.21) \& (8.56) \& (41.18) \& (0.0) \& (30.28) \& (2.56) \& (1.12)\\
\end{deptext}
\depedge{2}{1}{det}
\depedge{2}{5}{nmod}
\depedge{5}{3}{case}
\depedge{5}{4}{nummod}
\depedge{7}{2}{nsubjpass}
\depedge{7}{6}{auxpass}
\depedge{7}{8}{punct}
\end{dependency}&
\begin{tikzpicture}[node distance=1cm, auto, scale=.7, transform shape]
 \node[punkt, color=white] (input) {};
 \node[punkt, above=.1cm of input, color=gray] (gcn1) {GCN};
  \node[punkt, above=.4cm of gcn1, color=gray] (gcn2) {GCN};
 \node[punkt, above=.4cm of gcn2] (poolfc) {FC};
 \draw[->, thick,  color=red!70] (poolfc) -- (gcn2);
\end{tikzpicture}\\
\begin{dependency}[scale=.7, transform shape]
\begin{deptext}
|[top color=red!1.3599999999999999]|A \& |[top color=red!6.29]|total \& |[top color=red!4.58]|of \& |[top color=red!25.71]|116 \& |[top color=red!19.46]|patients \& |[top color=red!22.41]|were \& |[top color=red!19.13]|randomized \& |[top color=red!1.0699999999999998]|.\\
(1.36) \& (6.29) \& (4.58) \& (25.71) \& (19.46) \& (22.41) \& (19.13) \& (1.07)\\
\end{deptext}
\depedge[line width=0.609pt]{2}{1}{det}
\depedge[line width=0.501pt]{2}{5}{nmod}
\depedge[line width=1.098pt]{5}{3}{case}
\depedge[line width=2.82pt]{5}{4}{nummod}
\depedge[line width=1.796pt]{7}{2}{nsubjpass}
\depedge[line width=1.681pt]{7}{6}{auxpass}
\depedge[line width=0.508pt]{7}{8}{punct}
\end{dependency}& 
\begin{tikzpicture}[node distance=1cm, auto, scale=.7, transform shape]
 \node[punkt, color=white] (input) {};
 \node[punkt, above=.1cm of input, color=gray] (gcn1) {GCN};
  \node[punkt, above=.4cm of gcn1] (gcn2) {GCN};
 \node[punkt, above=.4cm of gcn2, color=gray] (poolfc) {FC};
 \draw[->, thick,  color=red!70] (gcn2) -- (gcn1);
\end{tikzpicture}\\
\begin{dependency}[scale=.7, transform shape]
\begin{deptext}
|[top color=red!0.74]|A \& |[top color=red!12.64]|total \& |[top color=red!1.15]|of \& |[top color=red!18.41]|116 \& |[top color=red!20.87]|patients \& |[top color=red!17.91]|were \& |[top color=red!27.389999999999997]|randomized \& |[top color=red!0.8999999999999999]|.\\
(0.74) \& (12.64) \& (1.15) \& (18.41) \& (20.87) \& (17.91) \& (27.39) \& (0.9)\\
\end{deptext}
\depedge[line width=0.593pt]{2}{1}{det}
\depedge[line width=1.898pt]{2}{5}{nmod}
\depedge[line width=1.014pt]{5}{3}{case}
\depedge[line width=1.595pt]{5}{4}{nummod}
\depedge[line width=1.038pt]{7}{2}{nsubjpass}
\depedge[line width=1.175pt]{7}{6}{auxpass}
\depedge[line width=0.526pt]{7}{8}{punct}
\end{dependency}&
\begin{tikzpicture}[node distance=1cm, auto, scale=.7, transform shape]
 \node[punkt, color=white] (input) {};
 \node[punkt, above=.4cm of input] (gcn1) {GCN};
  \node[punkt, above=.4cm of gcn1, color=gray] (gcn2) {GCN};
 \node[punkt, above=.4cm of gcn2, color=gray] (poolfc) {FC};
 \draw[->, thick, color=red!70] (gcn1) -- (input);
\end{tikzpicture}\\
\end{tabular}
\caption{Layerwise Relevance Visualization in a Graph Convolutional Network. Left: Projection of relevance percentages (in brackets) onto the input graph structure (red highlighting). Edge strength is proportional to the relevance percentage an edge carried from one layer to the next. Right: Architecture (replicated at each layer) of the GCN sentence classifier (input bottom, output top). Node and edge relevance were normalized layerwise. The predicted label of the input was \texttt{RESULT}, as was the true label.}
\label{fig:lrv}
\end{figure*}
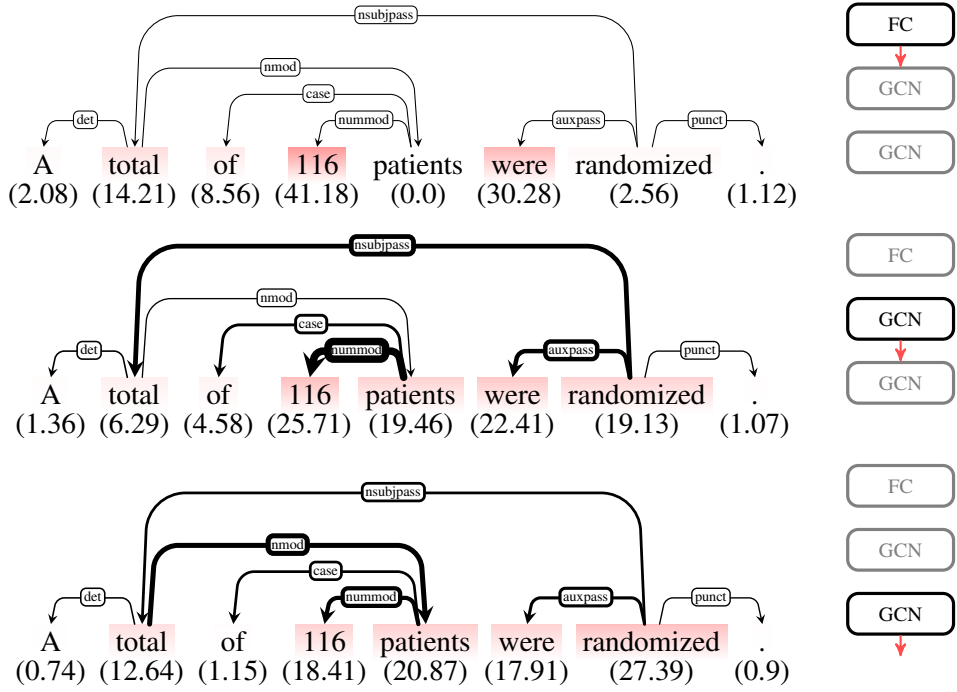
\section{Introduction}
A Deep Neural Network (DNN) that offers -- or from which we can retrieve -- explanations for its decisions arguably is easier to improve and debug than a black box model. Loosely following \citet{MonDSP18}, we understand an explanation as a ``collection of features (...) that have contributed for a given example to produce a decision.'' For humans to make sense of an explanation it needs to be expressed in a human-interpretable domain, which is often the input space \cite{MonDSP18}. 

For instance, in the vision domain \citet{bach-plos15} propose the Layerwise Relevance Propagation (LRP) algorithm that propagates contributions from an output activation back to the first layer, the input pixel space. \citet{arras2017relevant} implement a similar strategy for NLP tasks, where contributions are propagated into the word vector space and then summed up over the word vector dimensions to create heatmaps in the plain text input space. 

In the course of the backpropagation of contributions, hidden layers in the DNN are visited, but rarely inspected.\footnote{Although, \citet{arras2017relevant} propose inspecting hidden layers with LRP as a possible future direction.} A major challenge when inspecting and interpreting hidden states lies in the fact that a projection into an interpretable domain is often made difficult by the non-linearity, non-locality and dimension reduction/expansion across hidden layers.

In this work, we present an explainability method that visits and projects visible and hidden states in Graph Convolutional Networks (GCN) \cite{kipf2016semi} onto the interpretable input space. Each GCN layer fuses neighborhoods in the input graph. For this fusion to work, GCNs need to replicate the input adjacency at each layer. This allows us to project their intermediate representations back onto the graph.

Our method, layerwise relevance visualization (LRV), not only visualizes static explanations in the hidden layers, but also tracks and visualizes hidden cross-layer dynamics, as illustrated in Fig.~\ref{fig:lrv}. In a qualitative and a quantitative analysis, we experimentally demonstrate that our approach yields meaningful explanations for a GCN sentence classifier.
\section{Methods}
In this section we explain how we combine GCNs and LRP for layerwise relevance visualization. We begin with a short recap of GCNs and LRP to establish a basis for our approach and to introduce notation.
 \subsection{Graph Convolutional Networks}

Assume a graph $G$ represented by $(\tilde{A}, H^{(0)})$ where $\tilde{A}$ is a degree-normalized adjacency matrix, which introduces self loops to the nodes in $G$, and $H^{(0)}\in {\rm I\!R}^{n \times d}$ an embedding matrix of the $n$ nodes in $G$, ordered in compliance with $\tilde{A}$. A GCN layer propagates such graph inputs according to 
\begin{equation}
\label{eq:gcn}
    H^{(l+1)} = \sigma \left(\tilde{A}H^{(l)}W^{(l)}\right),
\end{equation}
which can be decomposed into a feature projection $H'^{(l)}=H^{(l)}W^{(l)}$ and an adjacency projection $\tilde{A}H'^{(l)}$, followed by a non-linearity $\sigma$. 

The adjacency projection fuses each node with the features in its effective neighborhood. At each GCN layer, the effective neighborhood becomes one hop larger, starting with a one-hop neighborhood in the first layer. The last layer in a GCN classifier typically is fully connected (FC) and projects its inputs onto class probabilities. 
\subsection{Layerwise Relevance Propagation}
\label{sec:lrp}
To receive explanations for the classifications of a GCN classifier, we apply LRP. LRP explains a neuron's activation by computing how much each of its input neurons contributed\footnote{We use the terms \textit{contribution} and \textit{relevance} interchangeably.} to the activation. Contributions are propagated back layerwise.

\citet{montavon2017explaining} show that the positive contribution of neuron $h^{(l)}_{i}$, as defined in \citet{bach-plos15} with the $z^{+}$-rule, is equivalent to
\begin{equation}
\label{eq:lrp}
    R_{i} = \sum_{j}\frac{h_{i}^{(l)}w_{ij}^{+}}{\sum_{k}h^{(l)}_{k}w_{kj}^{+}}R_{j},
\end{equation}
where $\sum_{k}$ sums over the neurons in layer $l$ and $\sum_{j}$ over the neurons in layer $(l+1)$. Eq.~\ref{eq:lrp} only allows positive inputs, which each layer receives if the previous layers are activated using ReLUs.\footnote{LRP is capable of tracking negative contributions, too. In this work, we focus on positive contributions.} LRP has an important property, namely the relevance conservation property: $\sum_{j} R_{j \leftarrow k} = R_{k}, R_{j} = \sum_{k} R_{j \leftarrow k}$, which not only conserves relevance from neuron to neuron but also from layer to layer  \cite{bach-plos15}.
\subsection{Layerwise Relevance Visualization}
\label{sec:lrv}
We combine graph convolutional networks and layerwise relevance propagation for layerwise relevance visualization as follows. During training, GCNs receive input graphs in the form of $(\tilde{A}, H^{(0)})$ tuples. This is efficient and allows batching but it poses a problem for LRP: If the adjacency matrix is considered part of the input, one could argue that it should receive relevance mass. This, however, would make it hard to meet the conservation property.

Instead, we treat $\tilde{A}$ as part of the model in a post-hoc explanation phase, in which we construct an FC layer with $\tilde{A}$ as its weights. The GCN layer then consists of two FC sublayers. During the forward pass, the first one performs the feature projection and the second one -- the newly constructed one -- the adjacency projection. 

To make use of Eq.~\ref{eq:lrp} in the adjacency layer, we need to avoid propagating negative activations. This is why we apply the ReLU activation early, right after the feature projection, which yields the propagation rule
\begin{equation}
    \label{eq:xgcnforward}
    H^{(l+1)} = \tilde{A}\sigma(H^{(l)}W^{(l)}).
\end{equation}
When unrolled, this effectively just moves one adjacency projection from inside the first ReLU to outside the last ReLU. Alternatively, it should be possible to first perform the adjacency projection and afterwards the feature projection. 

Eq.~\ref{eq:xgcnforward} allows us to directly apply  Eq.~\ref{eq:lrp} to the two projection sublayers in a GCN layer. During LRP, we cache the intermediate contribution maps $R^{(l)} \in {\rm I\!R}^{n \times f}$ that we receive right after we propagated past the feature projection sublayer and compute node $i$'s contribution in that layer as
$R(i^{(l)}) = \sum_{c}R^{(l)}_{ic}$. In addition, we also compute the edge relevance $e^{(l)}_{(i,j)}$ between nodes $i$ and $j$ as the amount of relevance that it carried from layer $l-1$ to layer $l$. 

In what follows, we use $R(i^{(l)})$ and $e^{(l)}_{ij}$ to visualize hidden state dynamics (i.e. relevance flow) in the input graph. Furthermore, similar to \citet{mingming}, we make use of perturbation experiments to verify that our method indeed identifies the relevant components in the input graph.

\section{Experiments}
\label{sec:experiments}
Our experiments are publicly available: \url{https://github.com/DFKI-NLP/lrv/}. We trained a GCN sentence classifier on a 20k subset of the PubMed 200k RCT dataset  \cite{dernoncourt2017pubmed}. The dataset contains scientific abstracts in which each sentence is annotated with either of the 5 labels \textit{BACKGROUND}, \textit{OBJECTIVE}, \textit{METHOD}, \textit{RESULT}, or \textit{CONCLUSION}. \citet{dernoncourt2017pubmed} describe the task as a sequential classification task since all sentences in one abstract are classified in sequence, but we treat it as a conventional sentence classification task, classifying each sentence in isolation.

In a preprocessing step, we used ScispaCy \cite{Neumann2019ScispaCyFA} for tokenization and dependency parsing to generate the input graphs for the GCN sentence classifier. For the node embeddings in the dependency graph, we utilized pre-trained fastText embeddings \cite{mikolov2018advances}. Edge type information was discarded. 

Our classifier, implemented and trained using PyTorch \cite{paszke2017automatic}, consists of two stacked GCN layers (Eq.~\ref{eq:xgcnforward}), followed by a max-pooling operation and a subsequent FC layer. All layers were trained without bias and optimized with the Adam optimizer \cite{kingma2014adam}. After each optimization step we clamped the negative weights of the last FC layer to avoid negative outputs.

During training, we batched inputs, but in the post-hoc explanation phase, single graphs from the test set were forwarded through the network to implement the strategy outlined in Sec.~\ref{sec:lrv}. We first constructed the adjacency layer with $\tilde{A}$, then performed a forward pass during which we cached the inputs at each layer for LRP. 

LRP started at the output neuron with the maximal activation, before the softmax normalization. For all intermediate layers we used Eq.~\ref{eq:lrp}. Since coefficients in the input word vectors are negative, we used a special propagation rule \cite{montavon2017explaining}, which we omit here, due to space contraints. We cached the intermediate contribution maps right after the max-pooling operation, after the second and after the first (input space) feature projection layers in the GCN layers. 

For the qualitative analysis, we visualized the contributions of each node at each layer as well as the relevance flow over the graph's edges across layers, as outlined in Sec.~\ref{sec:lrv}. For the quantitative validation, we deleted a growing portion of the globally most relevant edges (starting with the most relevant ones) and monitored the model's classification performance. In a second experiment, we did the same starting with the least relevant edges. Here, \textit{global edge relevance} refers to the sum of an edge's relevance across all layers, $\sum_{l}e_{ij}^{(l)}$.
\section{Results}
After training, our model achieved a weighted F$_{1}$ score of 0.822 on the official (20k) test set. We then performed the post-hoc explanation with the trained model, receiving layerwise explanations as the one depicted in Fig.~\ref{fig:lrv}. More examples can be found in the appendix.

Fig.~\ref{fig:lrv} (top) shows which vectors in the final layer the GCN bases its classification decision on. The middle and bottom segments reveal how much relevant information the model fused in these vectors from neighboring nodes during the forward pass: 

According to the LRV in Fig.~\ref{fig:lrv}, the model primarily bases its classification decision on the fusion of the vector representations of \texttt{total}, \texttt{patients}, and \texttt{116}. The vector representations of these nodes are accumulated in the vector of \texttt{116} in two steps.

Interestingly, the vector of \texttt{patients} contributes a larger portion after it has been fused with \texttt{total}. Furthermore, \texttt{randomized} becomes more relevant after the first GCN layer, contributing to the second most relevant n-gram \texttt{(were, randomized)}. Other n-grams, such as \texttt{(A, total)} or \texttt{(of, patients)} appear less relevant. 

Note that if we had simply redistributed the contributions in the final layer, using only the adjacency matrix, without considering the activations in the forward pass as LRP does, these n-grams would have received an unsubstantiated share. Furthermore, a conventional, single explanation in the input space would have hardly revealed the hidden dynamics discussed above.

\begin{figure}[!tb]
\pgfplotstableread{top.dat}{\top}
\pgfplotstableread{bottom.dat}{\bottom}
\pgfplotsset{every axis legend/.append style={at={(0.05,0.05)}, anchor=south west}} 
\begin{tikzpicture}[scale=.9, transform shape]
\begin{axis}[minor tick num=1,
xlabel=Percentage of deleted edges, ylabel=Weighted F$_1$, legend style={draw=none}]
\addplot [dashed, black,very thick] table [x={per}, y={f}] {\bottom};
\addplot [black,very thick] table [x={per}, y={f}] {\top};
\legend{least relevant, most relevant}
\end{axis}
\end{tikzpicture}
\caption{Perturbation experiment.}
\label{fig:pertubation}
\end{figure}
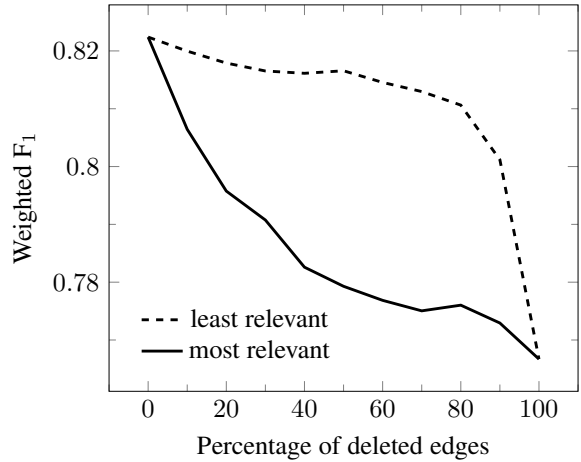

As explained in Sec.~\ref{sec:experiments}, we also conducted input perturbation experiments to validate that our method indeed identifies the graph components that are relevant to the GCN decision. Fig.~\ref{fig:pertubation} summarizes the results: The performance of our model degrades much faster when we delete edges that contributed a lot to the model's decision, according to our method. We take this as evidence that our method indeed correctly identifies the relevant components in the input graphs. 
\section{Related Work}
\citet{niepert2016learning, Wu-sgc2019} introduce graph networks that natively support feature visualization (explanations) but their models are not graph convolutional networks in the sense of \citet{kipf2016semi} that we address here.

\citet{velivckovic2017graph} propose Graph Attention Networks and also project an explanation of a hidden representation onto the input graph. They do not, however, continue across multiple layers. Furthermore, their explanation technique differs from ours. The authors use attention coefficients to scale edge thicknesses in their explanatory visualization. 
In contrast, we base edge relevance on the amount of relevance an edge carried from one layer to the next, according to LRP. Our method does this in a conventional GCN, which does not apply the attention mechanism. 



The body of work on the explainability of graph neural nets also includes the GNN explainer by \citet{ying2019gnn}, a model-agnostic explainability method. Since their method is model-agnostic, it can be applied to GCNs. Nevertheless, it would not reveal hidden dynamics, as our method does. 

In parallel to our work, in the last couple of months, several more works on explainability in the context of graph neural nets were published. \citet{pope2019explainability}, for instance, evaluate several explainability methods on data sets from the chemistry and vision domains. The methods do not include LRP, however, and the authors do not visualize relevance layerwise. The two works that are most related to our approach were published very recently by \citet{mingming} and \citet{baldassarre2019explainability}. Both implement LRP for graph networks (inter alia). 
\citet{mingming} introduce the notion of node importance visualization, which is also reflected in our explanations and \citet{baldassarre2019explainability}, similar to our approach,  suggest to trace (and visualize) explanations over feature-less edges. Both works, however, address a different task from ours: In contrast to our graph classification task, they perform node classification and identify k-hop neighbor contributions in the proximity of a central node of interest. 
\section{Conclusions}
We presented layerwise relevance visualization in convolutional text graph classifiers. We demonstrated that our approach allows to track, visualize and inspect visible as well as hidden state dynamics. We conducted qualitative and quantitative experiments to validate the proposed method. 

This is a focused contribution; further research should be conducted to test the approach in other domains and on other data sets. Future versions should also exploit LRP's ability to expose negative evidence and the method could be extended to node classifiers.
\section{Acknowledgements}
This research was partially supported by the German Federal Ministry of Education and Research through the project DEEPLEE (01IW17001). We would also like to thank the anonymous reviewers for their feedback on the paper and Arne Binder for his feedback on the code base. 
\bibliographystyle{acl_natbib}
\bibliography{emnlp-ijcnlp}
\pagebreak
\onecolumn
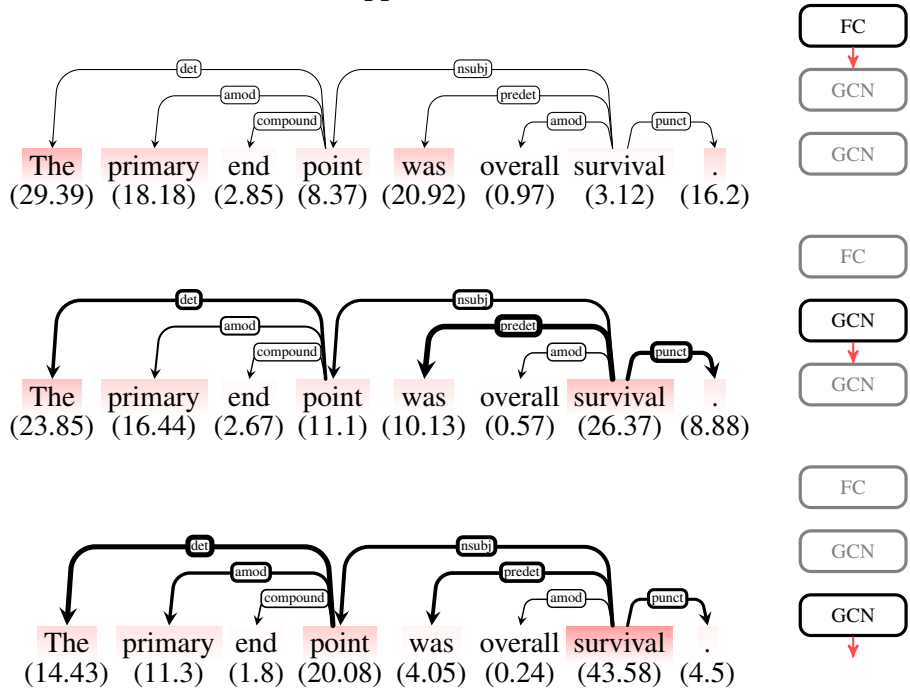
\begin{figure*}[!htb]

\textbf{A\quad Supplemental Material}
\centering
\begin{tabular}{cc}
\begin{dependency}[scale=.7, transform shape]
\begin{deptext}
|[top color=red!29.39]|The \& |[top color=red!18.18]|primary \& |[top color=red!2.85]|end \& |[top color=red!8.37]|point \& |[top color=red!20.919999999999998]|was \& |[top color=red!0.97]|overall \& |[top color=red!3.1199999999999997]|survival \& |[top color=red!16.2]|.\\
(29.39) \& (18.18) \& (2.85) \& (8.37) \& (20.92) \& (0.97) \& (3.12) \& (16.2)\\
\end{deptext}
\depedge{4}{1}{det}
\depedge{4}{2}{amod}
\depedge{4}{3}{compound}
\depedge{7}{4}{nsubj}
\depedge{7}{5}{predet}
\depedge{7}{6}{amod}
\depedge{7}{8}{punct}
\end{dependency} &
\begin{tikzpicture}[node distance=1cm, auto, scale=.7, transform shape]
 \node[punkt, color=white] (input) {};
 \node[punkt, above=.1cm of input, color=gray] (gcn1) {GCN};
  \node[punkt, above=.4cm of gcn1, color=gray] (gcn2) {GCN};
 \node[punkt, above=.4cm of gcn2] (poolfc) {FC};
 \draw[->, thick,  color=red!70] (poolfc) -- (gcn2);
\end{tikzpicture}\\
\begin{dependency}[scale=.7, transform shape]
\begin{deptext}
|[top color=red!23.849999999999998]|The \& |[top color=red!16.439999999999998]|primary \& |[top color=red!2.67]|end \& |[top color=red!11.1]|point \& |[top color=red!10.13]|was \& |[top color=red!0.5700000000000001]|overall \& |[top color=red!26.369999999999997]|survival \& |[top color=red!8.88]|.\\
(23.85) \& (16.44) \& (2.67) \& (11.1) \& (10.13) \& (0.57) \& (26.37) \& (8.88)\\
\end{deptext}
\depedge[line width=1.33pt]{4}{1}{det}
\depedge[line width=0.761pt]{4}{2}{amod}
\depedge[line width=0.527pt]{4}{3}{compound}
\depedge[line width=1.209pt]{7}{4}{nsubj}
\depedge[line width=2.119pt]{7}{5}{predet}
\depedge[line width=0.56pt]{7}{6}{amod}
\depedge[line width=1.598pt]{7}{8}{punct}
\end{dependency} &
\begin{tikzpicture}[node distance=1cm, auto, scale=.7, transform shape]
 \node[punkt, color=white] (input) {};
 \node[punkt, above=.1cm of input, color=gray] (gcn1) {GCN};
  \node[punkt, above=.4cm of gcn1] (gcn2) {GCN};
 \node[punkt, above=.4cm of gcn2, color=gray] (poolfc) {FC};

 \draw[->, thick,  color=red!70] (gcn2) -- (gcn1);
\end{tikzpicture}\\
\begin{dependency}[scale=.7, transform shape]
\begin{deptext}
|[top color=red!14.430000000000001]|The \& |[top color=red!11.3]|primary \& |[top color=red!1.7999999999999998]|end \& |[top color=red!20.080000000000002]|point \& |[top color=red!4.05]|was \& |[top color=red!0.24]|overall \& |[top color=red!43.580000000000005]|survival \& |[top color=red!4.5]|.\\
(14.43) \& (11.3) \& (1.8) \& (20.08) \& (4.05) \& (0.24) \& (43.58) \& (4.5)\\
\end{deptext}
\depedge[line width=1.913pt]{4}{1}{det}
\depedge[line width=1.271pt]{4}{2}{amod}
\depedge[line width=0.631pt]{4}{3}{compound}
\depedge[line width=1.467pt]{7}{4}{nsubj}
\depedge[line width=1.412pt]{7}{5}{predet}
\depedge[line width=0.55pt]{7}{6}{amod}
\depedge[line width=1.157pt]{7}{8}{punct}
\end{dependency} &
\begin{tikzpicture}[node distance=1cm, auto, scale=.7, transform shape]
 \node[punkt, color=white] (input) {};
 \node[punkt, above=.4cm of input] (gcn1) {GCN};
  \node[punkt, above=.4cm of gcn1, color=gray] (gcn2) {GCN};
 \node[punkt, above=.4cm of gcn2, color=gray] (poolfc) {FC};

 \draw[->, thick, color=red!70] (gcn1) -- (input);
\end{tikzpicture}\\
\end{tabular}
\caption{Predicted label: \texttt{METHOD}, true label: \texttt{METHOD}.}
\end{figure*}
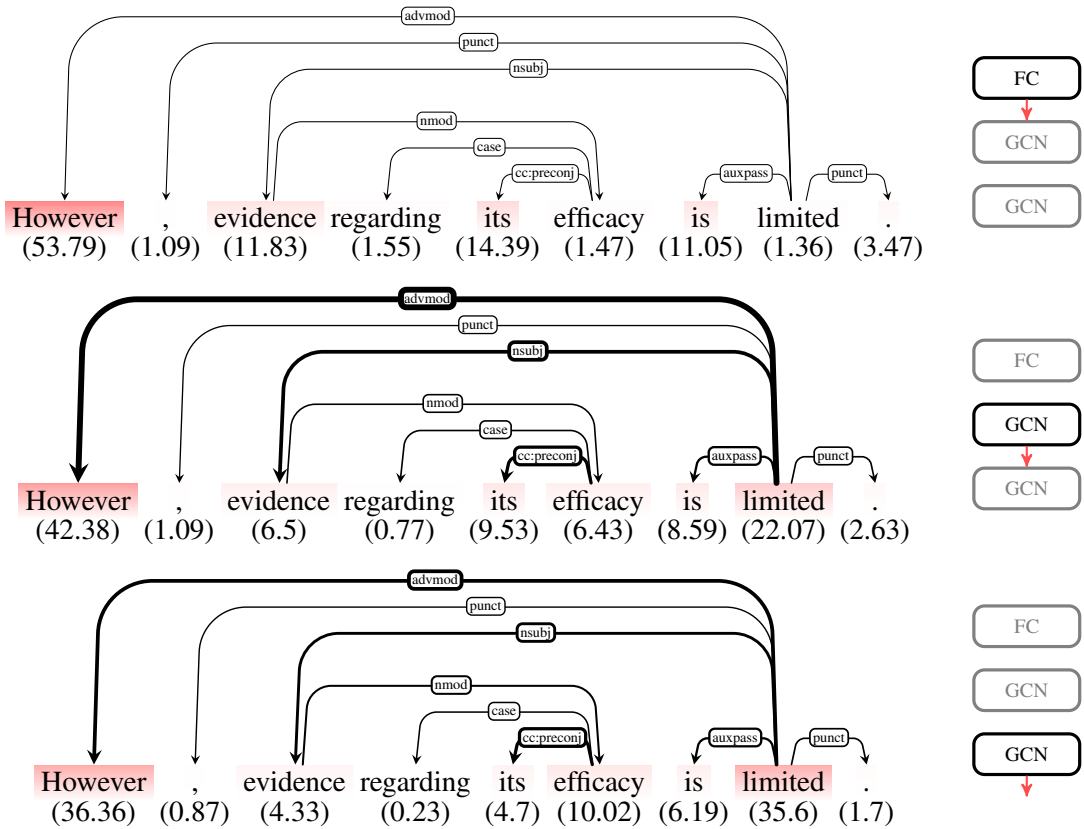
\begin{figure*}[!htb]

\centering
\begin{tabular}{cc}
\begin{dependency}[scale=.7, transform shape]
\begin{deptext}
|[top color=red!53.790000000000006]|However \& |[top color=red!1.09]|, \& |[top color=red!11.83]|evidence \& |[top color=red!1.55]|regarding \& |[top color=red!14.39]|its \& |[top color=red!1.47]|efficacy \& |[top color=red!11.05]|is \& |[top color=red!1.3599999999999999]|limited \& |[top color=red!3.47]|.\\
(53.79) \& (1.09) \& (11.83) \& (1.55) \& (14.39) \& (1.47) \& (11.05) \& (1.36) \& (3.47)\\
\end{deptext}
\depedge{3}{6}{nmod}
\depedge{6}{4}{case}
\depedge{6}{5}{cc:preconj}
\depedge{8}{1}{advmod}
\depedge{8}{2}{punct}
\depedge{8}{3}{nsubj}
\depedge{8}{7}{auxpass}
\depedge{8}{9}{punct}
\end{dependency}&
\begin{tikzpicture}[node distance=1cm, auto, scale=.7, transform shape]
 \node[punkt, color=white] (input) {};
 \node[punkt, above=.1cm of input, color=gray] (gcn1) {GCN};
  \node[punkt, above=.4cm of gcn1, color=gray] (gcn2) {GCN};
 \node[punkt, above=.4cm of gcn2] (poolfc) {FC};
 \draw[->, thick,  color=red!70] (poolfc) -- (gcn2);
\end{tikzpicture}\\
\begin{dependency}[scale=.7, transform shape]
\begin{deptext}
|[top color=red!42.38]|However \& |[top color=red!1.09]|, \& |[top color=red!6.5]|evidence \& |[top color=red!0.77]|regarding \& |[top color=red!9.53]|its \& |[top color=red!6.43]|efficacy \& |[top color=red!8.59]|is \& |[top color=red!22.07]|limited \& |[top color=red!2.63]|.\\
(42.38) \& (1.09) \& (6.5) \& (0.77) \& (9.53) \& (6.43) \& (8.59) \& (22.07) \& (2.63)\\
\end{deptext}
\depedge[line width=0.603pt]{3}{6}{nmod}
\depedge[line width=0.617pt]{6}{4}{case}
\depedge[line width=1.229pt]{6}{5}{cc:preconj}
\depedge[line width=2.211pt]{8}{1}{advmod}
\depedge[line width=0.5pt]{8}{2}{punct}
\depedge[line width=1.402pt]{8}{3}{nsubj}
\depedge[line width=0.869pt]{8}{7}{auxpass}
\depedge[line width=0.626pt]{8}{9}{punct}
\end{dependency}&
\begin{tikzpicture}[node distance=1cm, auto, scale=.7, transform shape]
 \node[punkt, color=white] (input) {};
 \node[punkt, above=.1cm of input, color=gray] (gcn1) {GCN};
  \node[punkt, above=.4cm of gcn1] (gcn2) {GCN};
 \node[punkt, above=.4cm of gcn2, color=gray] (poolfc) {FC};
 \draw[->, thick,  color=red!70] (gcn2) -- (gcn1);
\end{tikzpicture}\\
\begin{dependency}[scale=.7, transform shape]
\begin{deptext}
|[top color=red!36.36]|However \& |[top color=red!0.8699999999999999]|, \& |[top color=red!4.33]|evidence \& |[top color=red!0.22999999999999998]|regarding \& |[top color=red!4.7]|its \& |[top color=red!10.02]|efficacy \& |[top color=red!6.1899999999999995]|is \& |[top color=red!35.6]|limited \& |[top color=red!1.7000000000000002]|.\\
(36.36) \& (0.87) \& (4.33) \& (0.23) \& (4.7) \& (10.02) \& (6.19) \& (35.6) \& (1.7)\\
\end{deptext}
\depedge[line width=0.767pt]{3}{6}{nmod}
\depedge[line width=0.581pt]{6}{4}{case}
\depedge[line width=1.224pt]{6}{5}{cc:preconj}
\depedge[line width=1.403pt]{8}{1}{advmod}
\depedge[line width=0.533pt]{8}{2}{punct}
\depedge[line width=1.092pt]{8}{3}{nsubj}
\depedge[line width=0.86pt]{8}{7}{auxpass}
\depedge[line width=0.64pt]{8}{9}{punct}
\end{dependency} &
\begin{tikzpicture}[node distance=1cm, auto, scale=.7, transform shape]
 \node[punkt, color=white] (input) {};
 \node[punkt, above=.4cm of input] (gcn1) {GCN};
  \node[punkt, above=.4cm of gcn1, color=gray] (gcn2) {GCN};
 \node[punkt, above=.4cm of gcn2, color=gray] (poolfc) {FC};
 \draw[->, thick, color=red!70] (gcn1) -- (input);
\end{tikzpicture}\\
\end{tabular}
\caption{Predicted label: \texttt{BACKGROUND}, true label: \texttt{OBJECTIVE}.}
\end{figure*}
\end{document}